# Zero Shot Context-Based Object Segmentation using SLIP (SAM+CLIP)

Arushi Arora[1*], Saaketh Koundinya[1*], Shreya Agarwal[1*]

[1]Department of Electrical and Computer Engineering, New York University
aa10350@nyu.edu, sg7729@nyu.edu, sa6981@nyu.edu

## Abstract

We present SLIP (SAM+CLIP), an enhanced architecture for zero-shot object segmentation. SLIP combines the Segment Anything Model (SAM) (Kirillov et al. 2023) with the Contrastive Language-Image Pretraining (CLIP) (Radford et al. 2021). By incorporating text prompts into SAM using CLIP, SLIP enables object segmentation without prior training on specific classes or categories. We fine-tune CLIP on a Pokemon dataset, allowing it to learn meaningful image-text representations. SLIP demonstrates the ability to recognize and segment objects in images based on contextual information from text prompts, expanding the capabilities of SAM for versatile object segmentation.

Our experiments demonstrate the effectiveness of the SLIP architecture in segmenting objects in images based on textual cues. The integration of CLIP's text-image understanding capabilities into SAM expands the capabilities of the original architecture and enables more versatile and context-aware object segmentation.

The code is available at the following link

## Introduction

Object segmentation, the task of identifying and delineating objects within an image, is a fundamental problem in computer vision with numerous applications. Traditional approaches require training on large labelled datasets specific to each object class or category, making them limited in their generalizability (Chen et al. 2017) and (Long, Shelhamer, and Darrell 2015). However, recent advancements in deep learning have led to the development of models capable of zero-shot object segmentation, allowing for object recognition without prior training on specific classes (Zheng et al. 2021) and (Cha and Wang 2022).

This paper introduces SLIP (SAM+CLIP), a synergistic architecture that combines the strengths of the Segment Anything Model (SAM) and the Contrastive Language-Image Pretraining (CLIP) model. While SAM excels at object segmentation using images and masks, it lacks the ability to incorporate textual information. To address this limitation, we introduce CLIP, a model that learns joint representations of images and text through contrastive learning.

[*]These authors contributed equally.

By fine-tuning CLIP, we train it to understand the contextual information present in the images. This trained CLIP model allows us to leverage text prompts as additional input to the segmentation process in SAM, enabling zero-shot object segmentation based on the extracted context.

The key advantage of SLIP is its ability to segment objects in any class or category without the need for specific training on each class. Instead of relying solely on image-based information, SLIP combines the visual understanding of CLIP with the segmentation capabilities of SAM, utilizing text prompts to guide the segmentation process. This novel integration of text prompts and image segmentation opens up new possibilities for object recognition and segmentation in a wide range of applications.

It is important to emphasize that SLIP does not require training on the segmented images themselves. Instead, it utilizes the original images in our dataset, applying the same transformations and contextual understanding learned by CLIP. We then evaluate the performance of SLIP on a Pokemon dataset, demonstrating its ability to recognize and segment objects in images based on text prompts. The experimental results highlight the effectiveness of SLIP in achieving zero-shot object segmentation, paving the way for advanced applications in computer vision.

## Our Contributions

- Proposed a novel approach for zero-shot object segmentation using textual prompts by leveraging the synergy between SAM and CLIP.
- Extended the capabilities of the CLIP model by fine-tuning it on a Pokemon Dataset and integrating it with SAM for zero-shot object segmentation based on textual prompts.
- Conducted comprehensive evaluations and comparisons of Pokemon segmentation results using text prompts with both pretrained and fine-tuned CLIP models integrated with SAM. Additionally, established a ResNet classifier as a baseline for evaluating the effectiveness of the proposed approach.

## Literature Survey

The Contrastive Language-Image Pre-training (CLIP) model (Radford et al. 2021) is a major breakthrough in the field of computer vision and natural language processing developed by OpenAI in 2021. CLIP learns a joint representation space for images and text by leveraging a large and diverse dataset of paired images and textual descriptions. The model is based on a transformer architecture that encodes both the visual and textual modalities into a shared embedding space. This enables the model to perform cross-modal retrieval of images and text, and to reason about visual concepts through language. CLIP has achieved state-of-the-art results in various tasks, including image classification, object detection, and image retrieval, outperforming previous state-of-the-art models. In addition, the model's ability to perform cross-modal reasoning has led to innovative applications in fields such as image captioning, visual question answering, and even generative art.

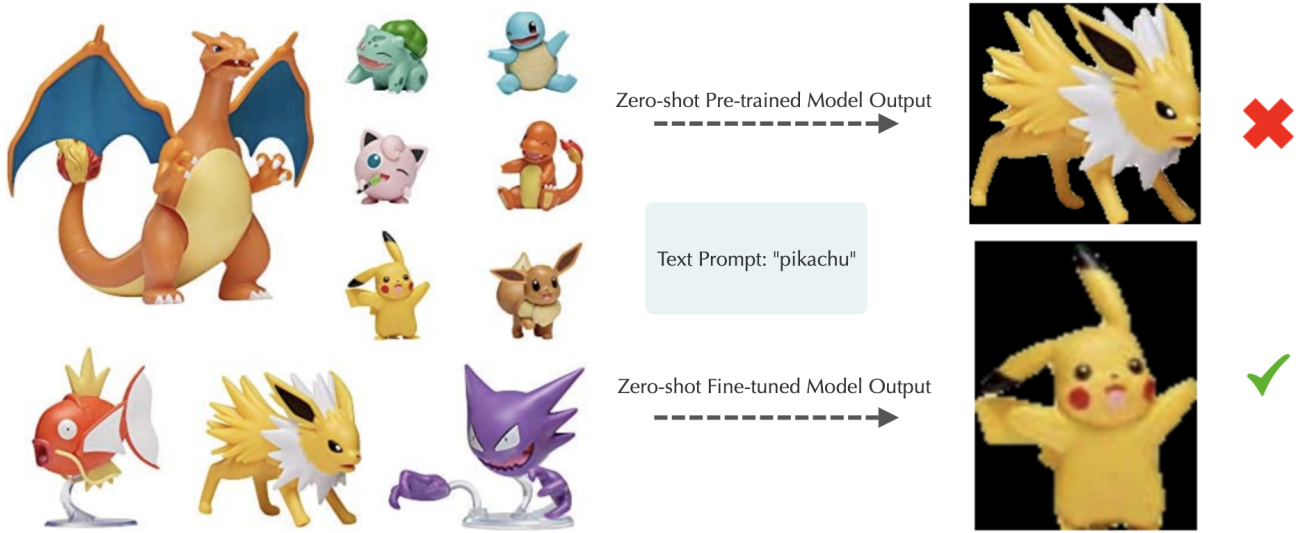

Figure 1: Output for Pretrained vs. Finetuned SLIP (SAM+CLIP)

Image segmentation is a fundamental problem in computer vision, which involves predicting a binary mask for each pixel in an image. Traditionally, this task has been tackled by training models on a fixed set of object classes, which can be costly to update or extend. In 2022, CLIPSeg (Lüddecke1 and Ecker1 2022) proposed a system that can generate image segmentations based on arbitrary prompts at test time, using a unified model trained once for multiple segmentation tasks. The system was built upon the CLIP model as a backbone and was extended with a transformer-based decoder for dense prediction. The binary segmentation map for an image was generated based on a free-text prompt or an additional image expressing the query. This novel hybrid input allowed dynamic adaptation not only to the three segmentation tasks but to any binary segmentation task where a text or image query can be formulated.

The Segment Anything model (Kirillov et al. 2023) is a revolutionary instance segmentation model developed by Meta Research and released in April 2023. The model was trained on 11 million images and 1.1 billion segmentation masks, making it the largest segmentation dataset to date. One of the notable features of SAM is its promptability, which allows it to transfer its segmentation capabilities to new image distributions and tasks, such as drawing boxes around objects of interest. Despite its impressive zero-shot performance on various segmentation tasks, SAM currently lacks the ability to incorporate text prompts into the segmentation process, indicating an opportunity for improvement in this area. The availability of the massive dataset, combined with the efficiency of the model, enables SAM to achieve strong zero-shot performance, often outperforming fully supervised approaches. These capabilities are particularly useful in scenarios where manual annotation of large datasets is not feasible, such as in medical imaging or remote sensing.

## Methodology

### Dataset

To effectively segment objects, which are not found in large-scale datasets like ImageNet (Deng et al. 2009) we finetuned CLIP on the Pokemon dataset (Dwivedi 2018). This helps us show how effective fine-tuning CLIP works in complex tasks such as object segmentation.

The Pokemon dataset is organized into 151 categories, corresponding to each Generation One Pokemon. Each category contains 50-60 images representing different perspectives, poses, and variations of the respective Pokemon. In total, the dataset consists of over 10,000 images, offering a comprehensive collection for training and evaluation purposes.

By utilizing this dataset, we can enhance the performance of models like SLIP (SAM+CLIP) for zero-shot object segmentation, enabling accurate and context-aware segmentation of Pokemon objects in images, even without prior training on specific Pokemon segments.

**Data Pre-processing**

During the data preprocessing phase, several steps were performed to prepare the dataset for further analysis and training. Since, the dataset consists only images, we have manually generated captions for each image. We have experimented with various caption generations like "This is an image of <pokemon name>", and "<pokemon name>". This enabled us to train CLIP effectively on image, text pairs, generating both positive and negative samples. We also found that the latter style of caption generation produced better contextual understanding over the former style.

**Model Architecture**

Our proposed model architecture combines the SAM (Segment Anything Model) and CLIP (Contrastive Language-Image Pretraining) frameworks to enable effective zero-shot object segmentation. The integration of these components leverages their respective strengths and enhances the overall performance of our model.

The SAM part of SLIP is responsible for generating accurate segments of all the objects present in the image and takes no contextual information. The process starts with passing an image to SAM's image encoder. The encoded image from the image encoder is then passed to the SAM Mask decoder. The SAM Mask decoder generates annotations corresponding to the input image. We have further improved these generated annotations using post-processing annotation filters. We created filters which separate out the intersecting segments without losing any relevant information. This step ensures that the generated segments accurately represent the objects present in the image and also improved the inference time of our model due to a significant reduction in segments.

Following the SAM Mask decoder, the segments are passed to the CLIP Image Encoder. Leveraging the finetuned CLIP model, the CLIP Image Encoder generates a similarity matrix that measures the similarity between the encoded image masks and the text prompt provided. This similarity matrix serves as a basis for selecting the segmented image that exhibits the highest similarity to the prompt, which is ultimately chosen as the model's output.

By synergistically combining the SAM image encoder, CLIP prompt encoder, SAM Mask decoder, and CLIP Image Encoder, SLIP harnesses the strengths of both frameworks to achieve robust zero-shot object segmentation. This seamless integration allows our model to generate highly accurate segmentations by effectively utilizing the contextual information provided by users through text prompts.

**Training**

To train our CLIP (Contrastive Language-Image Pretraining) model from scratch, we followed a series of steps and techniques. Here is an overview of the training process:

1. Dataloaders Preparation:

    We used the Pokemon dataset, consisting of approximately 10,000 images, for training our CLIP model. The dataset was split into an 80-20% train and validation split to evaluate the model's performance.

| Learning Rate | Projection Dimension | Validation Loss |
|---|---|---|
| 0.00001 | 128 | 0.9507 |
| 0.00001 | 512 | 0.9627 |
| **0.0001** | **128** | **0.7791** |
| 0.0001 | 512 | 0.7935 |
| 0.001 | 128 | 2.077 |
| 0.001 | 512 | 2.075 |
| 0.01 | 128 | 2.074 |
| 0.01 | 512 | 2.074 |
| 0.1 | 128 | 2.074 |
| 0.1 | 512 | 2.074 |
| 0.5 | 128 | 2.074 |
| 0.5 | 512 | 2.074 |

Table 1: Experiments with different Model Architectures for CLIP fine-tuning using grid search

2. Data Transformations:

    Prior to training, all images in the dataset were resized to a fixed size of 224x224 pixels with three color channels (224x224x3). Normalization techniques were applied to ensure consistent image representation across the dataset.

3. Pre-trained Image and Text Embeddings:

    We utilized pre-trained image embeddings trained on the ResNet-50 architecture, which captures rich visual features and representations from images. Similarly, pre-trained text embeddings trained on the DistilBERT model were used to encode textual information into meaningful representations.

4. Projection Head:

    The image embeddings and text embeddings from pre-trained models are projected into a common latent space using the projection head.

5. Logits and Targets:

    In order to train the model, we computed logits, which represent the similarity scores between the text and image embeddings. Ground truth similarity was used to generate targets, providing the model with the correct similarity values for each pair of text and image samples.

6. Cross-Entropy Loss:

    We calculated the cross-entropy loss between the logits and targets to measure the discrepancy between predicted similarities and ground truth similarities.

7. Optimization and Learning Rate Scheduling:

    We employed the AdamW optimizer to update the model's parameters during training. Additionally, a ReduceLROnPlateau scheduler was utilized to adjust the learning rate based on the model's performance. The learning rate was reduced by a factor of 0.9 after a certain number of epochs (patience=5).

8. Grid Search:

    To identify the optimal hyperparameters for training the CLIP model, we performed a grid search. We experimented with different learning rates (e.g., 0.00001, 0.0001, 0.001, 0.01, 0.1, 0.5) and projection dimensions

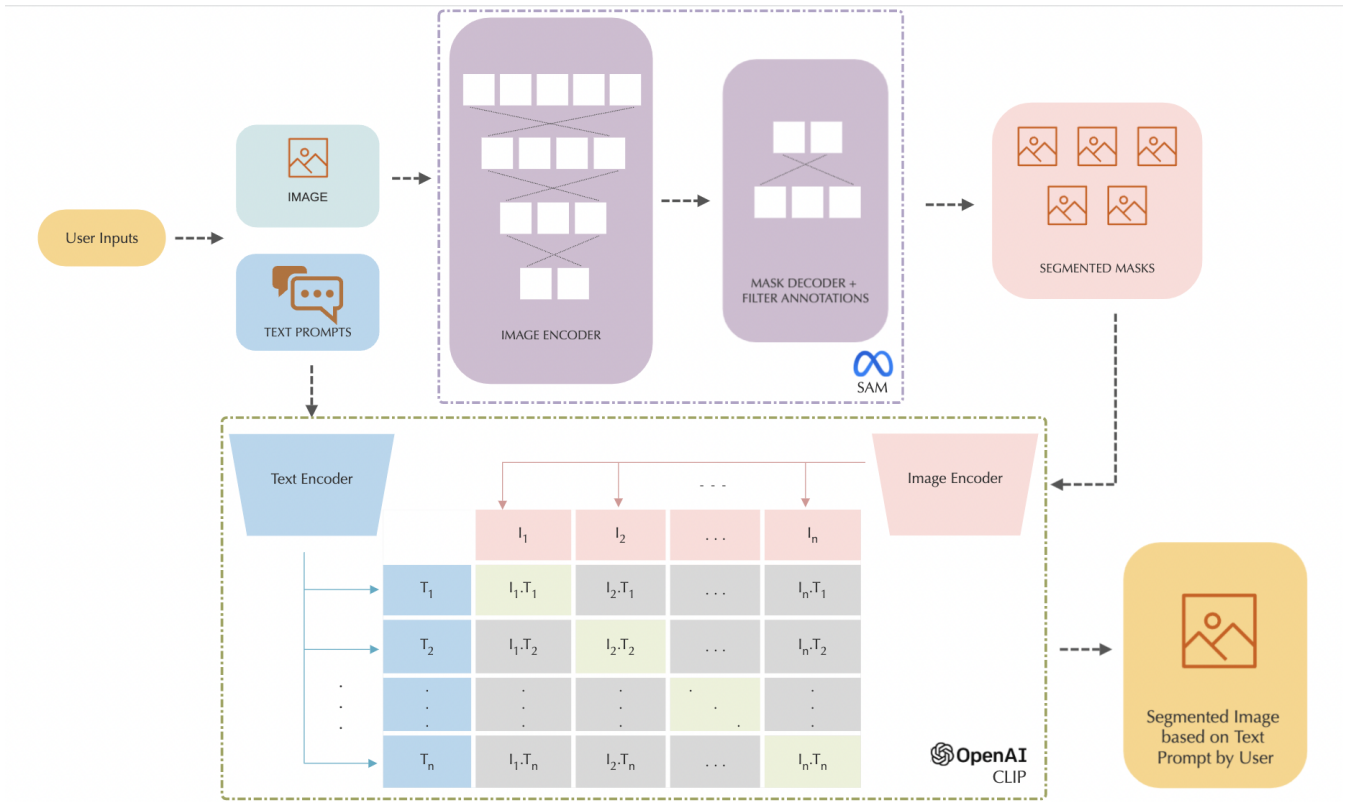

Figure 2: SLIP architecture overview: SLIP takes in an image and text prompt as input and generates a segment from the input image corresponding to the text prompt. SAM: takes image input and produces all possible segments, which are subsequently filtered. CLIP: takes the filtered segments and text prompt as input and outputs the segment which aligns closely with the input text.

(e.g., 128, 512) to find the combination that yielded the best results.

The training loss reached a value of 0.3710. The validation loss achieved a value of 0.7791, generalizing well on unseen images. The model was able to capture meaningful relationships between images and text, enabling it to accurately measure similarity and discriminate between different samples.

### Evaluation

In the process of evaluating our model, we analyzed its performance in predicting segments that are most closely related to the input prompt provided to the SLIP. Our evaluation was conducted on a dataset created by combining four random classes from the One Shot Pokemon Dataset (Yin 2018) into a single image, which served as the input to the SLIP along with an input prompt.

To accomplish this, we supplied the SLIP with both an input text prompt and an image containing the four combined classes. The model's output was the segment that exhibited the closest correspondence to the provided input. In order to determine the accuracy of these predictions, we established a baseline ResNet-18 model for image classification, which achieved an accuracy of 97%. This ResNet-18 model served as our ground truth for comparative purposes. If the label predicted by the classifier for the predicted segment matched the input text prompt, we considered the SLIP model's prediction of closest segment to the text prompt to be correct.

To further investigate the impact of training the CLIP model on our specific dataset, we conducted a comparative evaluation of the SLIP architecture both using pretrained CLIP and finetuned CLIP. This evaluation allowed us to gain insights into the improvements in performance resulting from training CLIP on our dataset, as well as its influence on the accuracy and robustness of segmentation and classification tasks.

It is important to note that our evaluation primarily focused on the accuracy of the model in providing the segment that is closest to the given input text prompt. The assessment of accuracy for actual segments falls within the scope of the SAM model, which is not the main focus of our current research. However, we suggest that in future studies, the SAM model could be fine-tuned using our dataset and considered for the purpose of segment evaluation.

### Results

The aim of our project was to enhance the capabilities of the Segment Anything Model (SAM) by incorporating text prompts for zero-shot object segmentation. SAM, although a powerful framework for object segmentation, lacked the ability to incorporate textual information as part of the segmentation process. Through our project, we successfully integrated the Contrastive Language-Image Pretraining (CLIP) model with SAM to enable text prompt-based object segmentation.

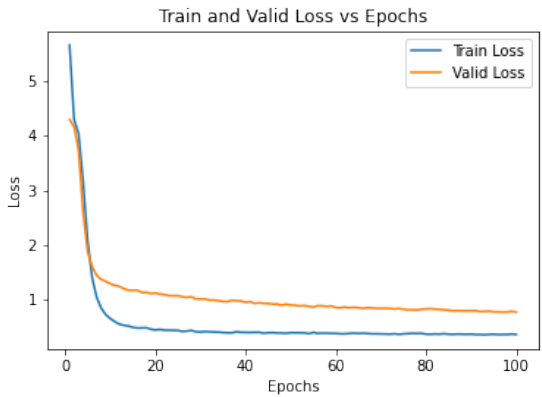

Figure 3: Training Progress of Trained CLIP at LR=0.0001

We trained the CLIP model on our specific dataset, which consisted of images of Pokemons. The training loss reached a value of 0.3710. The validation loss achieved a value of 0.7791, generalizing well on unseen images. The model was able to capture meaningful relationships between images and text, enabling it to accurately measure similarity and discriminate between different samples.

Using our proposed evaluation method and utilizing ResNet18 as baseline, the final accuracy of the fine-tuned SLIP model was 69.75%. The significant improvement in accuracy compared to the pre-trained SLIP model, which had an accuracy of 15.25%, demonstrates the effectiveness of our approach.

| Model | Accuracy |
|---|---|
| ResNet-18 (Supervised) | 97% |
| Pretrained SLIP | 15.25% |
| Finetuned SLIP | 69.75% |

Table 2: Results

## Conclusion

In conclusion, we introduce SLIP (SAM+CLIP), an advanced architecture that seamlessly integrates the Segment Anything Model (SAM) with the Contrastive Language-Image Pretraining (CLIP) model for zero-shot object segmentation.

By incorporating text prompts into SAM through CLIP, SLIP enables precise object segmentation without the need for pretraining on specific classes or categories. We specifically fine-tune CLIP on a comprehensive Pokemon dataset, enabling it to acquire rich image-text representations. Through extensive experimentation, we have demonstrated the effectiveness of the SLIP architecture in accurately recognizing and segmenting objects in images based on contextual information provided through text prompts. The integration of CLIP's text-image understanding capabilities significantly extends the capabilities of the original SAM architecture, enabling more versatile and context-aware object segmentation. The versatility and potential of the SLIP model extend to various applications, including image understanding, scene comprehension, and interactive image analysis, where the ability to perform object segmentation without prior class-specific training is immensely advantageous.

Overall, SLIP represents a significant advancement in zero-shot object segmentation, paving the way for innovative solutions in computer vision tasks.